\documentclass[10pt,twocolumn,letterpaper]{article}

\usepackage{cvpr}              

\usepackage{graphicx}
\usepackage{amsmath}
\usepackage{amssymb}
\usepackage{booktabs}

\usepackage[pagebackref,breaklinks,colorlinks]{hyperref}

\usepackage[capitalize]{cleveref}
\crefname{section}{Sec.}{Secs.}
\Crefname{section}{Section}{Sections}
\Crefname{table}{Table}{Tables}
\crefname{table}{Tab.}{Tabs.}

\def\cvprPaperID{*****} 
\def\confName{CVPR}
\def\confYear{2022}

\begin{document}

\title{NVIDIA-UNIBZ Submission for EPIC-KITCHENS-100 Action Anticipation Challenge 2022}

\author{Tsung-Ming Tai$^{1,2}$,
Oswald Lanz$^2$,
Giuseppe Fiameni$^1$,
Yi-Kwan Wong$^1$
\and
Sze-Sen Poon$^1$,
Cheng-Kuang Lee$^1$,
Ka-Chun Cheung$^1$,
Simon See$^1$\\[.1in]
$^1$NVIDIA, $^2$Free University of Bozen-Bolzano\\[.1in]
{\tt\small \{tstai,oswald.lanz\}@unibz.it}\\
{\tt\small \{gfiameni,gwong,spoon,ckl,chcheung,ssee\}@nvidia.com}
}
\maketitle

\begin{abstract}
In this report, we describe the technical details of our submission for the EPIC-Kitchen-100 action anticipation challenge.  Our modelings, the higher-order recurrent space-time transformer and the message-passing neural network with edge learning, are both recurrent-based architectures which observe only 2.5 seconds inference context to form the action anticipation prediction. By averaging the prediction scores from a set of models compiled with our proposed training pipeline, we achieved strong performance on the test set, which is 19.61\% overall mean top-5 recall, recorded as second place on the public leaderboard.
\end{abstract}

\section{Introduction}
\label{sec:intro}
Forecasting future events based on evidence of current conditions is an innate skill of human beings, and key for predicting the outcome of any decision making. Anticipating "\textit{what will happen next?}" is a natural skill for human beings, but not for machines. In computer vision, the same question arises in video action anticipation. It is a long standing and widely studied problem to recognize the human actions given a video clip. However, to further predict the future action based on the given observations has just attracted increasing interests in recent years. Unlike action recognition, in action anticipation, the target action only stays in causal relation to the signal in the sub-clip, but is not directly observable. It must be forecast as one possible consequence of the already observed video context. EPIC-Kitchen-100 \cite{damen2022rescaling} is the largest dataset containing the definition of the video action anticipation task. It considers 97 verbs and 300 nouns. Unique verb-noun pairs define 3807 action categories. The dataset is provided with the pre-extracted RGB, optical flow, object bounding box, and object mask modalities in this competition.

We participated in the video action anticipation challenge by considering two different proposed models:
\begin{itemize}
    \item \textit{Higher-Order Recurrent Space-Time Transformer} \cite{tai2021higher}: A recurrent network with space-time decomposition attention and higher order recurrent designs.
    \item \textit{Message-Passing Neural Network with Edge Learning} \cite{tai2022unified}: A recurrent network based on the message-passing framework. It models the sequential structure as a graph with a set of vertices and edges and learns the edge connectivity by different strategies. 
\end{itemize}

Both Higher-Order Recurrent Space-Time Transformer (HORST) and Message-Passing Neural Network with Edge Learning (MPNNEL) are recurrent architectures, and learn the spatial-temporal dependencies in different ways. HORST builds the n-gram temporal modeling by considering the higher-order recurrence with temporal attention and dynamically attends the relevant spatial information by spatial attention. On the other hand, MPNNEL projects the spatial contexts of frame input from each timestep onto the internal graph representation, and leverages the message-passing framework to capture the temporal propagation. MPNNEL also learns to augment the edge connectivity by using different end-to-end learning strategies. Both modelings are based on the extracted feature from 2D-CNN frame-based backbone. The final score for this competition was deployed by late-fusion of all the training variants from HORST and MPNNEL, and averaging the prediction scores of individual models across different modalities.   

The remaining parts of this report are organized as follows. The description of applied models is presented in Section 2, proposed training techniques are in Section 3. The experimental results are included in Section 4. Finally, Section 5 contains concluding remarks of this technical report.

\begin{figure*}
    \centering
    \includegraphics[width=1\textwidth]{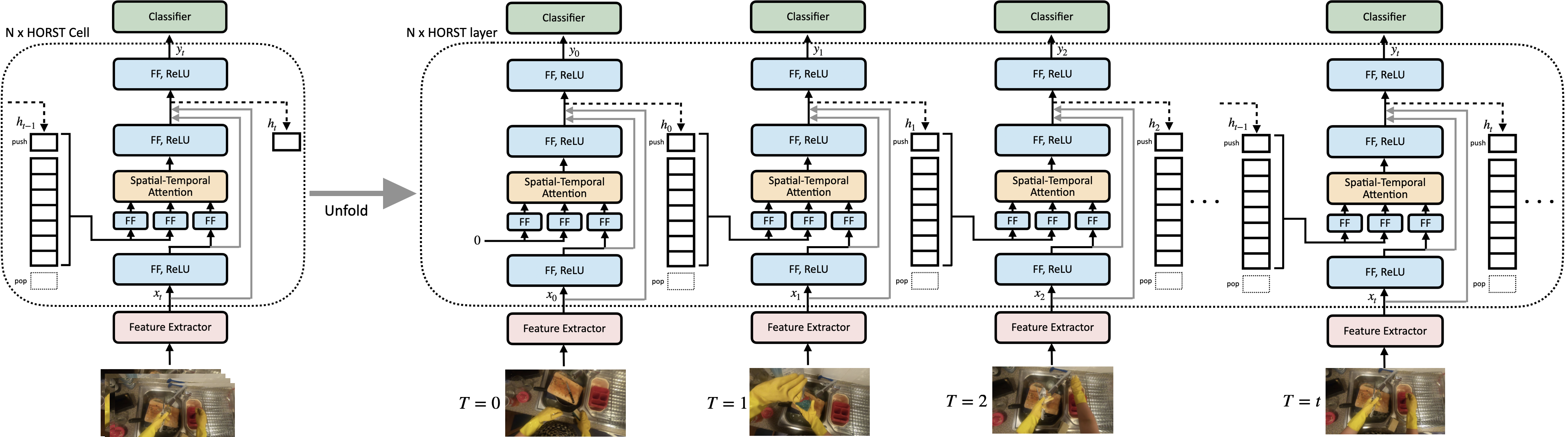}
    \caption{The overview HORST architecture. The HORST cell consist of a light-weighted spatial-temporal attention, and an internal first-in first-out queue to maintain the previous states for higher-order recurrence design.}
    \label{fig:horst_arch}
\end{figure*}

\begin{figure}
    \centering
    \includegraphics[width=.95\columnwidth]{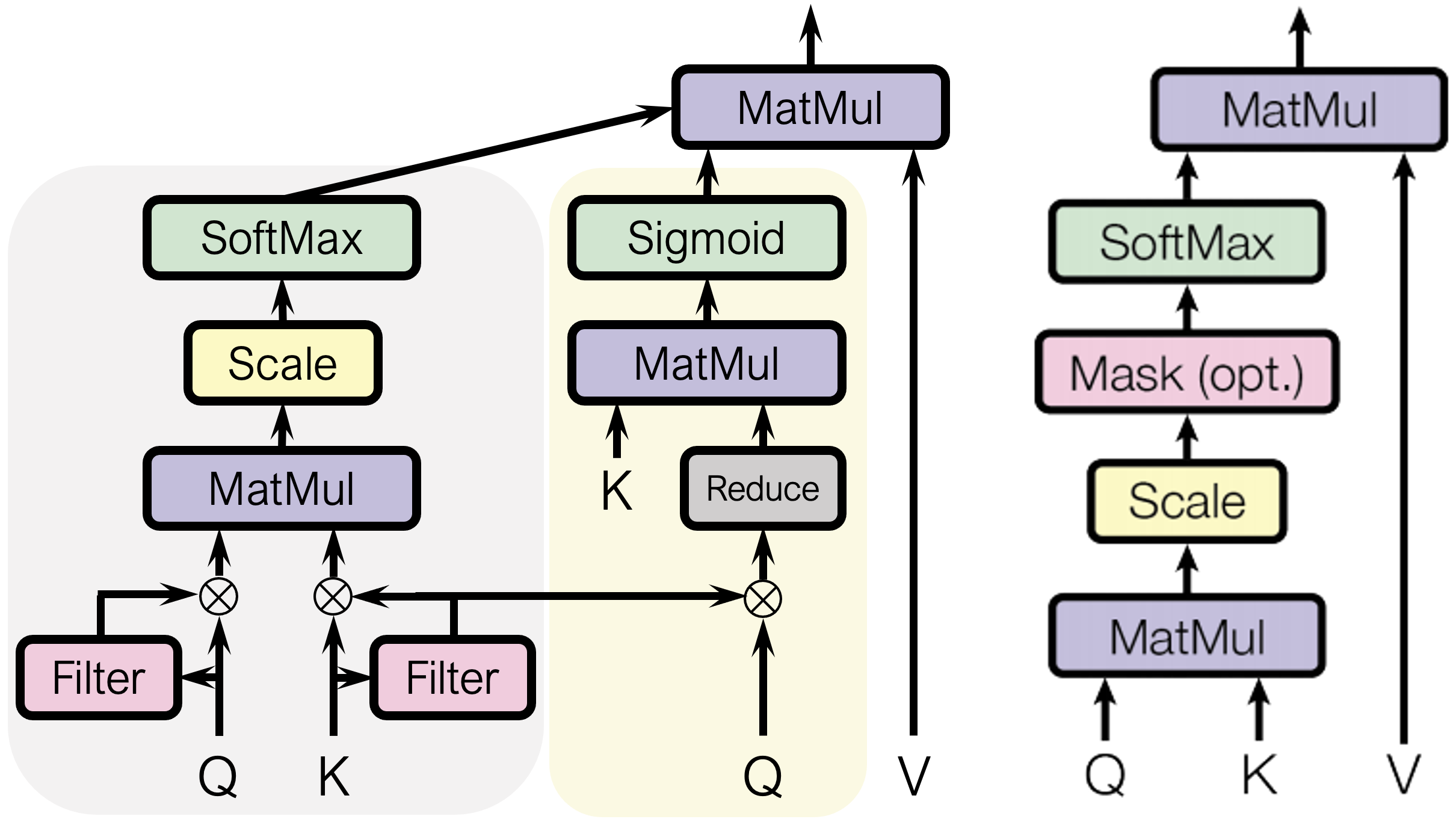}
    \caption{\textit{Left:} The space-time decomposition attention used in HORST; \textit{Right:} The self-attention proposed in \cite{vaswani2017attention}.}
    \label{fig:horst_statt}
\end{figure}

\section{Model Architecture}
\label{sec:architecture}
We briefly introduce HORST and MPNNEL architectures, the two modelings we used in this competition. 

\subsection{HORST Model}
To exploit the effective information in space-time structure, we proposed \textit{space-time decomposition attention} -- a light-weighted and computation-efficient attention, which integrates spatial and temporal operators from separated branches as shown in Figure~\ref{fig:horst_statt}. To define spatial and temporal branch operators, \textit{spatial filter} was introduced to recognize the relevant spatial information by the max and mean pooled features of inputs:
\begin{equation}
    f_{\mathcal{X}}(X) = \text{sigmoid}(\theta_{\mathcal{X}} * [X_{max}, X_{avg}] + b_{\mathcal{X}})
    \label{eq:filter}
\end{equation}
where $*$ is convolution, $X_{avg}, X_{max}$ are channel mean and max pooled, $\theta_\mathcal{X}$ and $b_{\mathcal{X}}$ are convolution kernels and biases.

The general higher-order recurrent network \cite{soltani2016higher,su2020convolutional,yu2017long} is with the following form:
\begin{equation}
    h_t = f(x_t, \phi(h_{t-1:t-S})),
\end{equation}
where the hidden state at time $t$, $h_t$, is computed by the cell function $f$ on input $x_t$ and $S$ orders states $h_{t-1:t-S}$ aggregated by the function $\phi$. 

The HORST cell can be viewed as instantiating $\phi$ with \textit{space-time decomposition attention} and maintain the previous states $h_{t-1:t-S}$ in an internal queue by the first-in first out update policy. The overall design is shown in Figure~\ref{fig:horst_arch}. At each step $t$, we process the video frame by a 2D-CNN backbone to obtain the feature map and encode it to the intermediate representation. Such representation is served as query and cross-reference from the historical states via the space-time decomposition attention. The attention output is then pushed to the queue while releasing the oldest state. Cell output finally propagate to the classifier. More details are found in \cite{tai2021higher} and we build our HORST models for this competition based on the codebase published at \url{https://github.com/CorcovadoMing/HORST}.

\begin{figure*}
    \centering
    \includegraphics[width=0.85\textwidth]{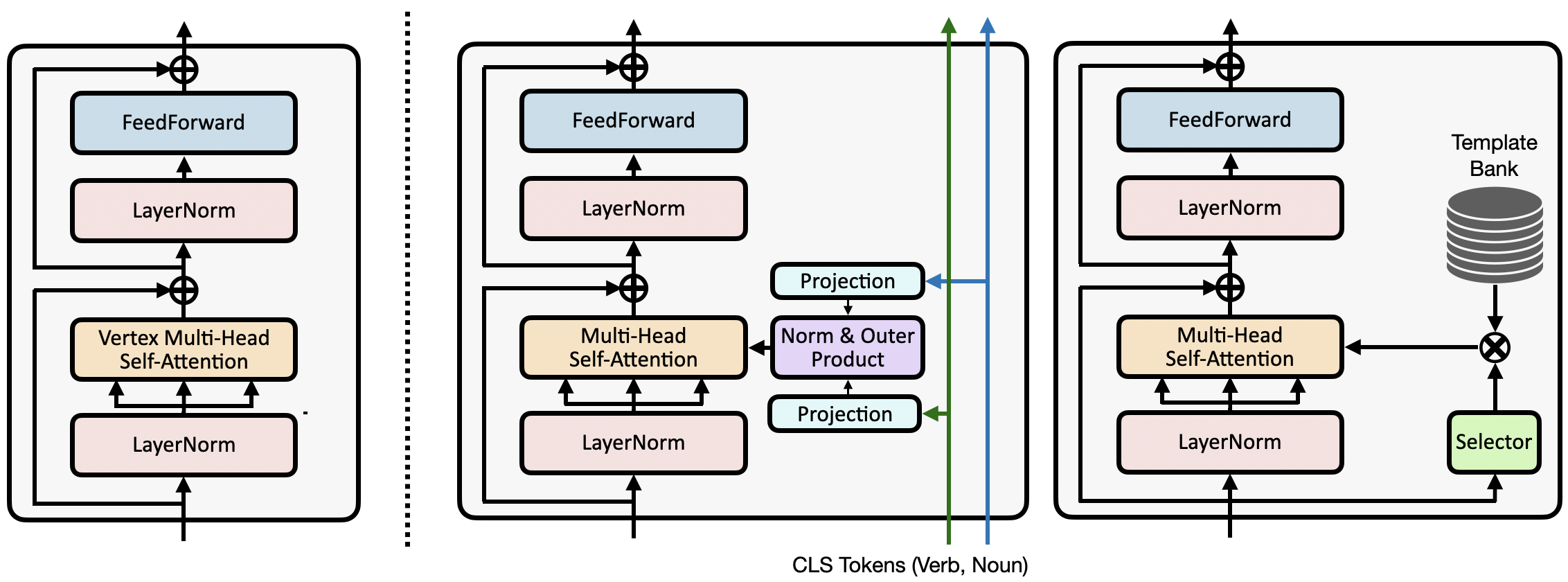}
    \caption{\textit{Left:} The implicit edge estimation by multi-head self-attention; \textit{Middle:} The augmented edge learning by outer product the class tokens supervised by the verb and noun annotations; \textit{Right:} The augmented edge learning by introducing a joint learnable template bank.}
    \label{fig:mpnn_edge_learning}
\end{figure*}

\begin{figure}
    \centering
    \includegraphics[width=1\columnwidth]{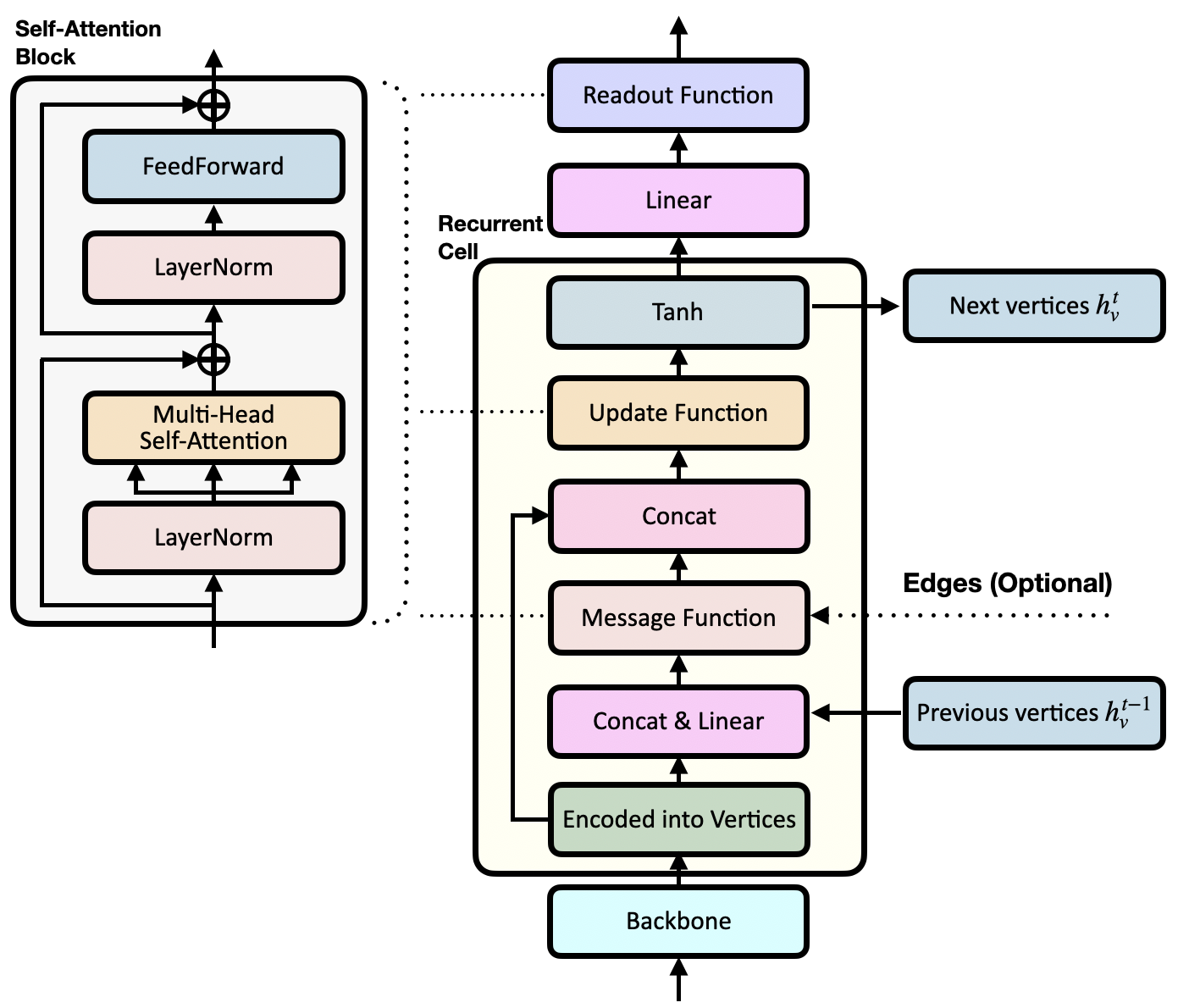}
    \caption{The overview architecture of MPNNEL. The message function is extendable with the explicit edge estimation by different edge learning strategies.}
    \label{fig:mpnn_arch}
\end{figure}

\subsection{MPNNEL Model}
MPNNEL translates the anticipation problem into a message passing scheme, producing a graph-structured space-time representation. The connectivity of the graph structure is inferred from the input at each time step. The readout function is called when the prediction is required at any timestep. The proposed model utilizes only multi-head self-attention for information routing between vertices. The overall architecture definition is illustrated in Figure~\ref{fig:mpnn_arch}. Note that the resulting spatial graph is either bi-directed, when an adjacency matrix $A$ is provided, or else it is un-directed. 

Without any prior knowledge, we assume each vertex in the graph can be accessible by any other vertices. In this case the scaled dot-product in the self-attention computes the pairwise similarity of all vertices from the inputs can be viewed as an \textit{implicit} edge estimation. This can be extended by optionally providing the edge estimation \textit{explicitly} by one of following strategies, also shown in Figure~\ref{fig:mpnn_edge_learning}:
\begin{itemize}
    \item \textit{Template Bank (TB)}, which forms the estimation of edge connections by soft-fusing a set of learnable templates using weights computed from the frame input.
    \item \textit{Class Token Projection (CTP)}, which performs the outer-product of class tokens to construct the edge estimation. The class tokens are supervised from provided verb and noun labels.
\end{itemize}

More details are found in \cite{tai2022unified} and we build our MPNNEL models for this competition based on the codebase published at \url{https://github.com/CorcovadoMing/MPNNEL}.

\section{Model Training}
\label{sec:training}
In this section, we describe the 4 phases training pipeline used to efficiently train all our models in this competition, and also the class weightings applied in the loss function to cope with imbalanced class distribution.

\subsection{Training Phases}
\begin{figure*}
    \centering
    \includegraphics[width=0.8\textwidth]{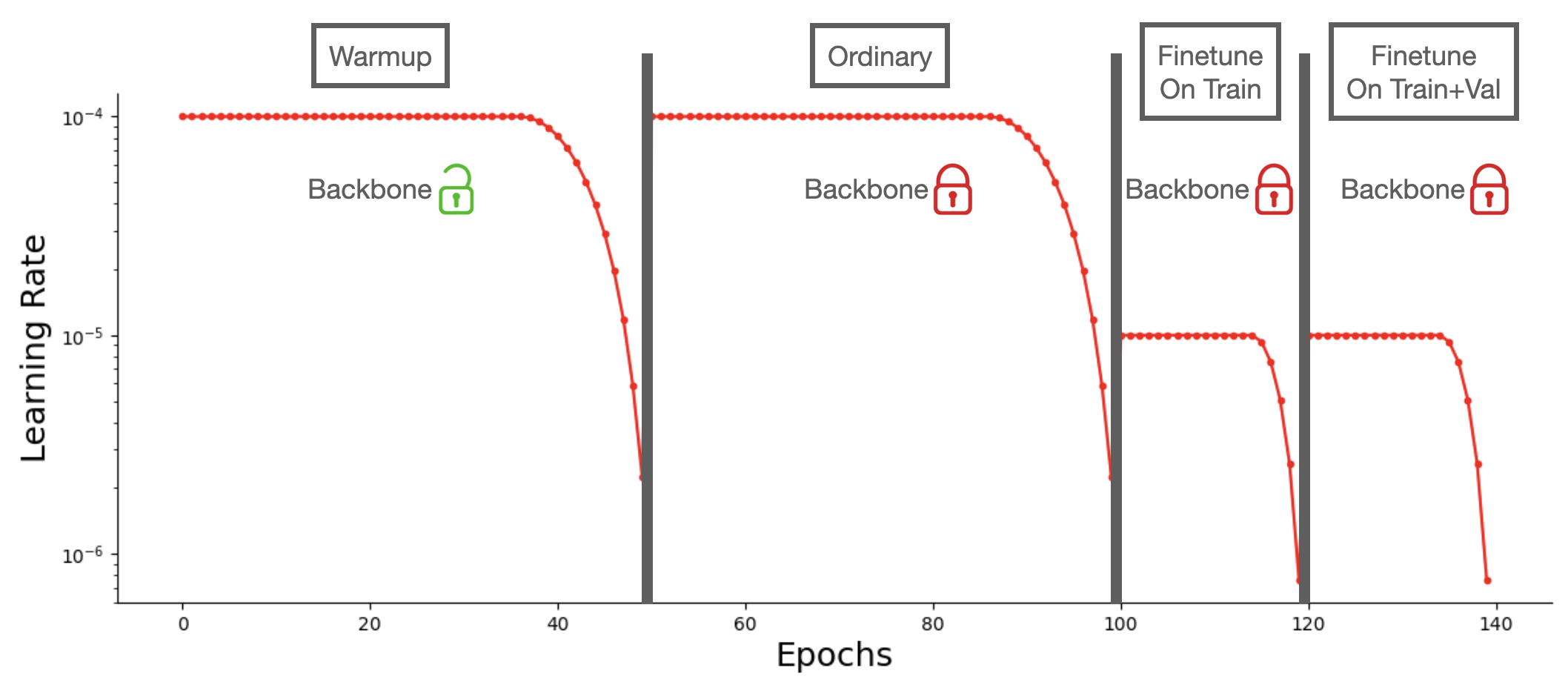}
    \caption{Demonstration of different training phase. The Backbone model is only trainable in the warmup phase and remains freeze in rest of the phases.}
    \label{fig:training_phase}
\end{figure*}
We trained all of our models by having them experience four learning phases, where they are (i) warmup phase; (ii) ordinary training; (iii) finetune; and (iv) finetune with joint validation set. The demonstration is shown in Figure~\ref{fig:training_phase}. 

The details of different training phases are:
\begin{itemize}
    \item \textit{Warmup Phase}: The model is end-to-end trainable on the target dataset. The model can access the actual action frames beyond the anticipation limitation only in this training phase.
    \item \textit{Ordinary Phase}: The model is trained with backbone freeze, and the action frames are not accessible in this and following training stages.
    \item \textit{Finetune Phase}: The model is trained with backbone freeze, under the lower learning rate, and with the class weightings adjusted in the loss functions.
    \item \textit{Finetune with joint validation set}: The model is trained with backbone freeze, under the lower learning rate, and with the class weightings adjusted. Additionally, the validation samples are joint together in the supervised learning.
\end{itemize}

The warmup phase is targeted to build a strong feature extractor for the competition, the backbone model is able to receive gradients and the action frames are allowed to be observable exclusively in this phase. The ordinary phase focus on the anticipation task and trains the HORST and MPNN architectures with the feature extractor kept frozen. The finetune phases learn to distinguish the hard samples and tailed cases by class weighting adjustments, also with the validation set jointly in the last training. Every phases are resumed from its previous step and each model training experiences the complete learning rate scheduling.

\subsection{Class Weightings}
We adjusted the class weightings of the cross-entropy loss for individual verbs, nouns, and actions during the fine-tuning stages. The adjustment is based on the label frequency summarized from the training set. Note the action distribution defined in EPIC-Kitchen-100 is composed of joint probability of verbs and nouns, however, the label frequency of the action class could be different than the individual frequency belonging to verbs and nouns. Therefore, we empirically found this adjustment brings additional regularization to the model learning and results in noticeable gains on the validation up to 4\% improvements.

\section{Experiments}
\label{sec:exp}
We provide implementation details and discuss the choices that led to our public record in the competition leaderboard. 

\subsection{Implementation Details}
We prepared each input modality as follow: RGB frames are resized to 224x224 and the pixel values are scaled from [0, 256] to [-1, 1]. The Flow modality came with the two maps described for horizontal and vertical optical changes, we stacked the two maps in channel dimension and resized them to 224x224 with pixel intensity scaled to [-1, 1]. As inspired from \cite{furnari2020rolling}, the Obj modality is formed by summarizing the object detection confidences from the officially provided object features, and discards the location information of the bounding boxes. Masked-RGB is the modality which multiplies the masking, extracted from a pretrained MaskedRCNN, with the RGB input.

The RandAugment \cite{cubuk2020randaugment} is applied for RGB, Masked-RGB, and Flow inputs. The video clip for training and inference are all sampled at 4 FPS (i.e., step 0.25s), as inherited from RU-LSTM baseline \cite{furnari2020rolling}. Each sample contains 14 sequential frames during training from 3.5s to 0.25s before action starts. However the last 3 frames are strictly not allowed to access in this competition since the anticipation time set to 1s. The total length of the inference context in our models are 2.5s (observed from 3.5s to 1s). 

We trained our model using batch size 32 on 4 $\times$ NVIDIA A100 GPUs. AdaBelief \cite{DBLP:conf/nips/ZhuangTDTDPD20} in combination with the look-ahead optimizer \cite{NEURIPS2019lookahead} is adopted. Weight decay is set to 0.001. The learning rate is set to 1e-4 and decreased to 1e-6 for warmup and ordinary training, and 1e-5 decreased to 1e-7 for finetune phases. The learning rate scheduling uses FlatCosine, which keeps the initial learning rate for the first 75\% of total epochs and switches to cosine schedule for the last 25\% epochs (see also Figure~\ref{fig:training_phase}). The total epochs for warmup and ordinary training are set to 50, and 20 for finetune phases.

\subsection{Individual Models}
\begin{table}
    \centering
    \small
    \caption{Individual model performance on validation set, measured in mean top-5 action recall (MT5R) at 1s, of various modalities using different modelings and backbones.}
    \begin{tabular}{l|l|l|c}
        Model & Modality & Backbone & MT5R (\%) \\
        \hline
        HORST & RGB & Swin-B & 18.42 \\
        HORST & RGB & ConvNeXt & 17.09 \\
        MPNNEL & RGB & Swin-B & 17.05 \\
        MPNNEL (CTP) & RGB & Swin-B & 18.18 \\
        MPNNEL (TB) & RGB & Swin-B & 17.05 \\
        MPNNEL & RGB & ConvNeXt & 17.18 \\
        MPNNEL (CTP) & RGB & ConvNeXt & 18.54 \\
        MPNNEL (TB) & RGB & ConvNeXt & 18.09 \\
        \hline
        HORST & Flow & Swin-B & 7.95 \\
        HORST & Flow & ConvNeXt & 7.36 \\
        HORST & Flow (Snippets) & Swin-B & 6.61 \\
        HORST & Flow (Snippets) & ConvNeXt & 8.06 \\
        MPNNEL & Flow & Swin-B & - \\
        MPNNEL (CTP) & Flow & Swin-B & 6.66 \\
        MPNNEL (TB) & Flow & Swin-B & - \\
        MPNNEL & Flow & ConvNeXt & 7.59 \\
        MPNNEL (CTP) & Flow & ConvNeXt & 8.74 \\
        MPNNEL (TB) & Flow & ConvNeXt & 8.18 \\
        \hline
        HORST & Obj & None & 8.72 \\
        MPNNEL & Obj & None & 9.69 \\
        MPNNEL (CTP) & Obj & None & 8.80 \\
        MPNNEL (TB) & Obj & None & 8.99 \\
        \hline
        HORST & Masked-RGB & Swin-B & 12.03 \\
        HORST & Masked-RGB & ConvNeXt & 11.30 \\
        MPNNEL & Masked-RGB & Swin-B & 9.22 \\
        MPNNEL (CTP) & Masked-RGB & Swin-B & 7.87 \\
        MPNNEL (TB) & Masked-RGB & Swin-B & 9.57 \\
        MPNNEL & Masked-RGB & ConvNeXt & 9.65 \\
        MPNNEL (CTP) & Masked-RGB & ConvNeXt & 8.53 \\
        MPNNEL (TB) & Masked-RGB & ConvNeXt & 10.30 \\
    \end{tabular}
    \label{tab:individual_models}
\end{table}
Unlike other modalities which are in an spatial-temporal structure, the Obj modality is presented as a temporal sequence of frame vectors. Each such vector represents the frame-level object scores computed from an object detection pretrained model. We modified HORST and MPNNEL models for supporting the 1D object vector representation, by replacing the 2D Convolution in HORST with the fully-connected layer; and by replacing the object entities with learnable vectors multiplied by corresponding object scores to defined the vertices in MPNNEL. Some models apply on Flow modality by snippets, as suggested in \cite{furnari2020rolling}, where the previous 5 sequential Flow features are stacked.

For all of our models we considered the Swin Transformer (i.e., base configuration, Swin-B) \cite{liu2021swin}, and ConvNeXt \cite{liu2022convnet} as backbones. We showed validation results of each representative category in Table~\ref{tab:individual_models}. Note the validation results reported in Table~\ref{tab:individual_models} are before training with the joint validation set, in order to keep the numbers meaningful.

\subsection{Model Ensemble}
\begin{table}
    \small
    \centering
    \caption{Test accuracy of model ensemble.}
    \begin{tabular}{l|c}
        Model & MT5R (\%) \\
        \hline
        (a) HORST Family with all modalities & 17.47 \\
        (b) MPNNEL Family with all modalities & 18.19 \\
        \hline
        (a) + (b) & 19.52 \\
        (a) + (b) and weightings 1.2x on all RGB models & \textbf{19.61}
    \end{tabular}
    \label{tab:model_ensmeble}
\end{table}
We manually selected the strong models from each individual variant, and tried to balance between HORST and MPNNEL instances to maintain the diversity among the ensembled models. Our best submission, 19.61\% overall accuracy, was achieved by an ensemble of in total 54 models. Those models consisted of  30 RGB models, 10 Flow models, 8 Obj models, and 6 Masked-RGB models.

Table~\ref{tab:model_ensmeble} reports on the trajectory we stepped to our highest score submission. Averaging the prediction scores in the HORST family resulted in 17.47\% overall test accuracy, and 18.19\% in the MPNNEL family. Combining both HORST and MPNNEL further improved the score significantly, to 19.52\%, indicating some degree of complementarity of the two recurrent models.  We also empirically found that emphasizing the prediction scores of all RGB models can have additional performance gains. In our best submission we weighted all RGB models by a factor 1.2x higher than other modalities.

\section{Conclusion}
In this report, we presented the technical details of our submission, achieving an overall 19.61\% mean top-5 recall on the EPIC-Kitchen-100 anticipation challenge 2022. Our method considered the Higher-Order Recurrent Space-Time Transformer (HORST) and Message-Passing Neural Network with Edge Learning (MPNNEL) architectures, which are both recurrent-based networks and only observed 2.5s inference context for the action anticipation. Combined with the proposed training pipeline and by averaging the prediction scores from the models trained from various modalities, our submission recorded the second place on the public leaderboard.

{\small
\bibliographystyle{ieee_fullname}
\bibliography{main}

\begin{thebibliography}{10}\itemsep=-1pt

\bibitem{cubuk2020randaugment}
Ekin~D Cubuk, Barret Zoph, Jonathon Shlens, and Quoc~V Le.
\newblock Randaugment: Practical automated data augmentation with a reduced
  search space.
\newblock In {\em Proceedings of the IEEE/CVF Conference on Computer Vision and
  Pattern Recognition Workshops}, pages 702--703, 2020.

\bibitem{damen2022rescaling}
Dima Damen, Hazel Doughty, Giovanni~Maria Farinella, Antonino Furnari,
  Evangelos Kazakos, Jian Ma, Davide Moltisanti, Jonathan Munro, Toby Perrett,
  Will Price, et~al.
\newblock Rescaling egocentric vision: Collection, pipeline and challenges for
  epic-kitchens-100.
\newblock {\em International Journal of Computer Vision}, 130(1):33--55, 2022.

\bibitem{furnari2020rolling}
Antonino Furnari and Giovanni~Maria Farinella.
\newblock Rolling-unrolling lstms for action anticipation from first-person
  video.
\newblock {\em IEEE Transactions on Pattern Analysis and Machine Intelligence},
  43(11):4021--4036, 2020.

\bibitem{liu2021swin}
Ze Liu, Yutong Lin, Yue Cao, Han Hu, Yixuan Wei, Zheng Zhang, Stephen Lin, and
  Baining Guo.
\newblock Swin transformer: Hierarchical vision transformer using shifted
  windows.
\newblock In {\em Proceedings of the IEEE/CVF International Conference on
  Computer Vision}, pages 10012--10022, 2021.

\bibitem{liu2022convnet}
Zhuang Liu, Hanzi Mao, Chao-Yuan Wu, Christoph Feichtenhofer, Trevor Darrell,
  and Saining Xie.
\newblock A convnet for the 2020s.
\newblock {\em arXiv:2201.03545}, 2022.

\bibitem{soltani2016higher}
Rohollah Soltani and Hui Jiang.
\newblock Higher order recurrent neural networks.
\newblock {\em arXiv:1605.00064}, 2016.

\bibitem{su2020convolutional}
Jiahao Su, Wonmin Byeon, Jean Kossaifi, Furong Huang, Jan Kautz, and Anima
  Anandkumar.
\newblock Convolutional tensor-train lstm for spatio-temporal learning.
\newblock In {\em Advances in Neural Information Processing Systems},
  volume~33, pages 13714--13726, 2020.

\bibitem{tai2021higher}
Tsung-Ming Tai, Giuseppe Fiameni, Cheng-Kuang Lee, and Oswald Lanz.
\newblock Higher order recurrent space-time transformer for video action
  prediction.
\newblock {\em arXiv:2104.08665}, 2021.

\bibitem{tai2022unified}
Tsung-Ming Tai, Giuseppe Fiameni, Cheng-Kuang Lee, Simon See, and Oswald Lanz.
\newblock Unified recurrence modeling for video action anticipation.
\newblock {\em arXiv:2206.01009}, 2022.

\bibitem{vaswani2017attention}
Ashish Vaswani, Noam Shazeer, Niki Parmar, Jakob Uszkoreit, Llion Jones,
  Aidan~N Gomez, {\L}ukasz Kaiser, and Illia Polosukhin.
\newblock Attention is all you need.
\newblock In {\em Advances in Neural Information Processing Systems},
  volume~30, 2017.

\bibitem{yu2017long}
Rose Yu, Stephan Zheng, Anima Anandkumar, and Yisong Yue.
\newblock Long-term forecasting using tensor-train rnns.
\newblock {\em arXiv:1711.00073}, 2017.

\bibitem{NEURIPS2019lookahead}
Michael Zhang, James Lucas, Jimmy Ba, and Geoffrey~E Hinton.
\newblock Lookahead optimizer: k steps forward, 1 step back.
\newblock In {\em NeurIPS}, volume~32, 2019.

\bibitem{DBLP:conf/nips/ZhuangTDTDPD20}
Juntang Zhuang, Tommy Tang, Yifan Ding, Sekhar~C. Tatikonda, Nicha~C. Dvornek,
  Xenophon Papademetris, and James~S. Duncan.
\newblock Adabelief optimizer: Adapting stepsizes by the belief in observed
  gradients.
\newblock In {\em NeurIPS}, 2020.

\end{thebibliography}
}

\end{document}